\documentclass[twoside]{article}

\usepackage{amsmath}
\usepackage{amsfonts}
\usepackage{algorithm}
\usepackage{algorithmic}
\usepackage{graphicx}
\usepackage{subcaption}
\usepackage{url}

\usepackage[accepted]{aistats2021}
\begin{document}



\twocolumn[

\aistatstitle{Gaussian Processes on Graphs via Spectral Kernel Learning}

\aistatsauthor{ Yin-Cong Zhi$^{\dagger}$ \And Yin Cheng Ng$^{\ddagger}$ \And Xiaowen Dong$^{\dagger}$}

\aistatsaddress{ $^{\dagger}$Department of Engineering Science, University of Oxford \\  $^{\ddagger}$Man AHL } ]

\newcommand{\E}{\mathbb{E}}
\newcommand{\Var}{\textmd{Var}}
\newcommand{\Cov}{\textmd{Cov}}
\newcommand{\p}{\mathbb{P}}
\newcommand{\N}{\mathcal{N}}
\newcommand{\f}{\mathbf{f}}
\newcommand{\B}{\mathbf{B}}

\begin{abstract}
We propose a graph spectrum-based Gaussian process for prediction of signals defined on nodes of the graph. The model is designed to capture various graph signal structures through a highly adaptive kernel that incorporates a flexible polynomial function in the graph spectral domain. Unlike most existing approaches, we propose to learn such a spectral kernel defined on a discrete space. In addition, this kernel has the interpretability of graph filtering achieved by a bespoke maximum likelihood learning algorithm that enforces the positivity of the spectrum. We demonstrate the interpretability of the model in synthetic experiments from which we show the various ground truth spectral filters can be accurately recovered, and the adaptability translates to superior performances in the prediction of real-world graph data of various characteristics.
\end{abstract}

\section{Introduction}

Graphs are highly useful data structures that represent relationships and interactions between entities. Such relational structures are commonly observed in the real-world, but can also be artificially constructed from data according to heuristics. The graph structure can be exploited in conjunction with other auxiliary data to build more powerful predictive models. One particular class of models that can be enhanced for graph data is Gaussian processes (GP). As a kernel method, GPs can be adapted to incorporate topological information through kernels derived on graphs. With the kernel defined, the standard Bayesian inference machinery can be directly applied to yield predictions.

Multi-output Gaussian processes (MOGP) are regression models for vector-valued data. Given a set of input covariates and the corresponding output vectors, the model makes vectorial predictions given a novel input covariate. In graph signal prediction problems, each output signal, indexed by a corresponding input, can be viewed as a vector where the dependency between elements is encoded in the graph structure. The dependency between the signals can then be modeled using a typical kernel on the inputs (e.g., the squared exponential kernel). The formulation of separable kernels for MOGP, as is the case in co-regionalization model in \cite{alvarez2012kernels}, makes choosing the overall kernel function straight forward. The two kernels operating on the inputs and output signals can be designed separately and combined by means of a Kronecker product.
% (as shown in equation (3)). 
We refer to the kernel operating on the input as kernel on the input space, and the kernel operating on the signals as kernel on the graph/output space, where the latter provides a measure of smoothness between data observed on the nodes.

Smola et al \cite{smola2003kernels} have introduced the notion of kernel on graphs, where kernel functions between nodes were derived from a regularization perspective by specifying a function in the graph spectral domain. The resulting kernel is based on the graph Laplacian, and this is closely related to graph signal processing, which makes use of tools such as graph Fourier transform and filtering \cite{shuman2012emerging, ortega2018graph, chen2014signal}. One particular low-pass filter defined in \cite{shuman2012emerging}, commonly used to de-noise graph signals, also assumes the form of kernels on graphs. This was subsequently used in % by Venkitaraman et al 
\cite{venkitaraman2018gaussian, venkitaraman2017kernel} for predicting low pass graph signals. However, not all data exhibits a low-pass spectrum, and this filter will not work as well on data of band- and high-pass spectra. The same limitation also applies to other existing GP models developed for graph-structured data such as \cite{ng2018bayesian}, where the relationship between the node observations is defined a priori. Addressing this limitation requires a different choice of kernels with a spectrum that better compliments the characteristics of the data.

Learning kernels in the spectral domain have been studied in the continuous case such as \cite{Wilson2013GaussianPK,samo2015generalized,li2019automated}, but the extension of the approach to learning on a discrete graph space has yet to be explored with the use of a GP. In this paper, we propose a MOGP model that uses a kernel on graphs for the output space, with the addition that the spectrum of the kernel is learnt. Our model constitutes several contributions to the literature: First, the model is designed to capture various graph signal structures by incorporating a flexible polynomial function in the graph spectral domain, producing a highly adaptable model. Second, The polynomial function is learnt by maximizing the log-marginal likelihood while respecting a constraint to enforce the positivity of the spectrum. The positivity constraint allows for a meaningful interpretation of the learnt models as graph filters, giving the modelers insights on the characteristics of the data. Finally, we demonstrate that our algorithm can recover ground truth filters applied to synthetic data, and show the adaptability of the model on real-world data with different spectral characteristics.

\section{Background}

\subsection{Gaussian Processes}

A GP $f$ is defined as
\begin{align}
    f(\mathbf{x}) \sim \mathcal{GP}\big(m(\mathbf{x}), \mathcal{K}(\mathbf{x},\mathbf{x'})\big)
\end{align}
for any inputs $\mathbf{x}, \mathbf{x'}$, where $m(\cdot)$ is the mean function, and $\mathcal{K}(\cdot,\cdot)$ is the symmetric and positive definite kernel function.
% The $\mathcal{K}$ belongs to the family of kernel functions with the properties of being symmetric and positive definite, and these form the essence of GP models.
In machine learning, GPs are widely employed 
% for assigning priors, tractable posteriors and 
models for making predictions with uncertainty. We will refer readers to \cite{rasmussen2003gaussian} for a more thorough description of the GP model.

\subsection{Spectral Filtering on Graphs}

Let $\mathcal{G}$ be a graph with vertex set $V$ such that $|V| = M$, we define the notion of spectral filtering on graphs from the graph Laplacian \cite{shuman2012emerging} defined as $\mathbf{L} = \mathbf{D} - \mathbf{A}$, where $\mathbf{A}$ is the adjacency matrix and $\mathbf{D}$ is the diagonal degree matrix. Assuming that $\mathcal{G}$ is undirected, the Laplacian admits the eigen-decomposition $\mathbf{L} = \mathbf{U\Lambda U}^\top$ where $\mathbf{U}$ contains the eigenvectors and $\mathbf{\Lambda}$ is the diagonal matrix of eigenvalues. A signal $\mathbf{y} \in \mathbb{R}^M$ on $\mathcal{G}$ can be viewed as a function
\begin{align}
    \mathbf{y}: V \rightarrow \mathbb{R},
\end{align}
and the graph Fourier transform of the signal, defined as $\mathbf{U^\top y}$, computes the spectrum of $\mathbf{y}$ to produce the amplitude of each eigenvector (frequency) component. Filtering then involves a function $g(\mathbf{\Lambda})$ in the graph spectral domain that may reduce or amplify each component leading to a filtered signal
\begin{align}
    \mathbf{U} g(\mathbf{\Lambda}) \mathbf{U^\top y}.
\end{align}
The term $\mathbf{U} g(\mathbf{\Lambda}) \mathbf{U}^\top$ is therefore referred to as a filtering function on the graph characterized by $g$.

\subsection{Kernels and Regularization on Graphs}

A property of kernel functions is provided by Bochner's theorem \cite{loomis2013introduction}, which states that positive definite functions have non-negative measures as the spectrum in the spectral domain. On the discrete graph space kernels are derived by the graph Fourier transform and a non-negative transfer function. In this section we briefly summarize the formulation of kernel on graphs described in \cite{smola2003kernels}.

The graph Laplacian can be used to quantify the smoothness of a function on graphs by measuring how much they vary locally \cite{smola1998regularization, smola2001regularization}. When finding a smooth model $\f$ for graph signal $\mathbf{y}$, it is common to solve for the following regularized problem
\begin{align}
    \min_{\f}||\f - \mathbf{y}||^2_2 + R(\mathbf{f}),
\end{align}
where we have the regularization function $R$ on $\f$. In the graph case, $R(\f) = \f^\top \mathbf{P} \f$ where $\mathbf{P}$ often takes the form of a penalty function of the graph Laplacian - $\mathbf{P} = r(\mathbf{L})$ - that penalizes specific graph spectral components of $\f$. The kernel function is then computed by $\mathbf{K} = \mathbf{P}^{-1}$, with pseudoinverse used if $\mathbf{P}$ is singular \cite{bozzo2013moore}. More generally, kernels on graphs assume the following form
\begin{align}
    \sum_{i=1}^M r^{-1}(\lambda_i) \mathbf{v}_i \mathbf{v}_i^\top =
    \mathbf{U}r^{-1}(\mathbf{\Lambda})\mathbf{U}^\top =  r^{-1}(\mathbf{L}) \label{kernel_on_graph}
\end{align}
for diagonal matrix $\Lambda$ containing the eigenvalues (i.e., $\{\lambda_i\}_{i=1}^M$) of $\mathbf{L}$ in increasing order and $\mathbf{U}$ containing the corresponding eigenvectors (i.e., $\{v_i\}_{i=1}^M$). Furthermore, this definition is flexible in that different variations of the Laplacian such as the normalized Laplacian 
\begin{align}
    \tilde{\mathbf{L}} = \mathbf{D}^{-\frac{1}{2}}\mathbf{LD}^{-\frac{1}{2}}
\end{align}
and scaled Laplacian
\begin{align}
    \mathbf{L}_S = \frac{1}{\lambda_{max}(\mathbf{L})}\mathbf{L} \label{scaled}
\end{align}
will both lead to valid kernels.

\section{Proposed Model}

%\subsection{Gaussian Process with Separable Kernel}
\subsection{Gaussian Processes for Graph Signals}
\label{sec:gpg}

From a generative model perspective, consider data pairs of the form $\{\mathbf{x}_n, \mathbf{y}_n\}_{n=1}^N$ where each $\mathbf{y}_n \in \mathbb{R}^M$ is a signals on a graph $\mathcal{G}$ of $M$ nodes. We assume each $\mathbf{y}_n$ is a realization of a filtering system $\B \f(\mathbf{x}_n)$ indexed by the covariate $\mathbf{x}_n$ where $\B \in \mathbb{R}^{M\times M}$ is the graph filter, and $\f \in \mathbb{R}^{M}$ is a simple MOGP function evaluated at $\mathbf{x}_n$ with independent components. The elements in $\f$ are assumed to be independent GPs with identical kernel function $\mathcal{K}$ on any two inputs $\mathbf{x}$ and $\mathbf{x}^\prime$. This leads to $\Cov(\f(\mathbf{x}_n), \f(\mathbf{x}_m)) = \mathcal{K}(\mathbf{x}_n, \mathbf{x}_m)\mathbf{I}_M$, where $\mathbf{I}_M \in \mathbb{R}^{M\times M}$ is an identity matrix. Graph information in $\mathbf{y}_n$ is therefore induced by the filtering matrix $\B$, giving rise to the following model
\begin{align}
    \mathbf{y}_n = \B\f(\mathbf{x}_n) + \boldsymbol{\epsilon}_n,
    \label{filtering}
    %\label{eq:gen}
\end{align}
where $\mathbf{\boldsymbol{\epsilon}}_n \sim \N(0, \sigma_\epsilon^2\mathbf{I}_M)$. The model in Eq. (\ref{filtering}) is generic in the sense that, depending on the design of $\B$, we can incorporate any characteristics of the signal $\mathbf{y}_n$ in the graph spectral domain.

The prior covariance between two signals $\mathbf{y}_n$ and $\mathbf{y}_m$ can be computed as $\Cov(\mathbf{y}_n, \mathbf{y}_m) = \E(\mathbf{y}_n \mathbf{y}_m^\top) = \B\E (\f(\mathbf{x}_n)\f(\mathbf{x}_m)^\top) \B^\top = \mathcal{K}(\mathbf{x}_n, \mathbf{x}_m) \B\B^\top$, and if we let $\tilde{\mathbf{y}} = \textmd{vec}(\mathbf{y}_1, \dots, \mathbf{y}_N)$, the covariance of the full data becomes
\begin{align}
    \Cov (\tilde{\mathbf{y}}) = \mathbf{K}\otimes \B\B^\top + \sigma^2_\epsilon \mathbf{I}_{MN}, \label{input_output_space}
\end{align}
where $\mathbf{K}_{nm} = \mathcal{K}(\mathbf{x}_n, \mathbf{x}_m)$, and $\otimes$ denotes the Kronecker product. The $\B\B^\top$ term can be thought of as a kernel between elements of each outputs $\mathbf{y}_n$, while $\mathbf{K}$ operates on signals' inputs $\mathbf{x}_n$ and $\mathbf{x}_m$. Generally, $\mathbf{K}$ will be referred to as the kernel on the input space, while we will call $\B\B^\top$ the graph dimension or the kernel on the output space.

We now state our main model for prediction of graph signals. Given the GP prior on the latent $\f(\mathbf{x})$, the training signals $\tilde{\mathbf{y}} = \text{vec}(\mathbf{y}_1, \dots, \mathbf{y}_N)$ and test signal $\mathbf{y}_* \in \mathbb{R}^M$ with given input $\mathbf{x}_*$ follow the joint distribution
\begin{align}
    \p \bigg( \begin{bmatrix}\tilde{\mathbf{y}} \\ \mathbf{y}_* \end{bmatrix} \bigg) \sim \N \bigg( \; 0 \; , \; \begin{bmatrix} \mathbf{K} \otimes \B\B^\top & \mathbf{K}_* \otimes \B\B^\top \\ \mathbf{K}_*^\top \otimes \B\B^\top & \mathbf{K}_{**} \otimes \B\B^\top \end{bmatrix} + \sigma_\epsilon^2 \mathbf{I}_{MN} \bigg). \label{multigau_local}
\end{align}
For the model covariances $\mathbf{K}_* = \mathcal{K}(\mathbf{x}_i, \mathbf{x}_*)_{i=1}^N \in \mathbb{R}^N$ and $\mathbf{K}_{**} = \mathcal{K}(x_*, x_*)$, where $\mathcal{K}$ can be any existing kernel such as the squared exponential or Matérn kernel. In this work, we consider $\B$ as a kernel on graphs based on the scaled graph Laplacian of eq. (\ref{scaled}), and follow the general form (\ref{kernel_on_graph}) as $\B = \sum_{i=1}^M g(\lambda_i) \mathbf{v}_i \mathbf{v}_i^\top = g(\mathbf{L}_S)$, where $\lambda_i$ and $\mathbf{v}_i$ from this point onwards correspond to the eigenvalues and eigenvectors of $\mathbf{L}_S$, and $g(\lambda)$ is the function in the graph spectral domain. The resulting $\B$ can be interpreted as both a filter and a kernel on the graph.

\subsection{Graph Spectral Kernel Learning}
\label{sec:skl}

To further enhance our model, we propose to learn the function $g$ rather than using an existing kernel on graphs to make the model adaptive. We parameterize $g$ as a finite polynomial function
\begin{align}
    g(\lambda) = \beta_0 + \beta_1 \lambda + \dots + \beta_P \lambda^P
    \implies \B = \sum_{i=0}^P \beta_i \mathbf{L}_S^i,
\end{align}
with coefficients $\beta_0, \dots, \beta_P$ learnt via log-marginal likelihood maximization. We will use gradient optimization to learn these coefficients which we will go into more details in the next section. A suitable choice for the degree can be found via a validation procedure; in practice, we find % the model does not require going beyond 
that a choice of $P = 3$ often leads to satisfactory performances.

There are a number of advantages of our model setup, in particular:
\begin{itemize}%[noitemsep]
    \item The kernel on graphs is learnt rather than chosen a priori, and the function that characterizes the kernel is a flexible polynomial making the model highly adaptable to data with different spectral properties. Moreover, existing choices provided in \cite{smola2003kernels} all consist of functions that have polynomial expansions. Hence our model provides suitable approximations if data came from a more complex generative model.
    \item The scaled Laplacian ensures the eigenvalues lie in the full range $[0,1]$ regardless of the graph. This bounds the values in the polynomial, and helps prevent the distribution of the eigenvalues from affecting the shape of $g$. Other alternatives such as the normalized Laplacian $\tilde{\mathbf{L}} = \mathbf{D^{-\frac{1}{2}}LD^{-\frac{1}{2}}}$ often found in the literature of graph signal processing \cite{shuman2012emerging} bounds the eigenvalues to be within $[0,2]$ and, by subtracting the identity matrix, shifts the eigenvalues to the range $[-1, 1]$. However, we find that often the eigenvalues are not spread over the full range $[-1, 1]$, thus the polynomial is only defined partially over the range.
    \item The application of the $P^{th}$ power of the Laplacian corresponds to filtering restricted to the $P$-hop neighbourhood of the nodes. Our polynomial is finite, thus the user can control the localization in the kernel, a property that is often desirable in graph based models such as the GCN \cite{kipf2016semi}.
    \item The linearity of $\beta_0, \dots, \beta_P$ means differentiation is efficient, making the function suitable for gradient optimization.
\end{itemize}

\subsection{Equivalence to the Co-regionalization Model}

The prior model in (\ref{multigau_local}) follows the form of separable kernels similar to the co-regionalization model in the literature of multi-output GP \cite{alvarez2012kernels}. Our derivation specifies the kernel on the output space more directly, but in this section we show how we can arrive at our model from the co-regionalization setup. Starting with the model $\mathbf{y}_n = \f(\mathbf{x}_n) + \mathbf{\epsilon}_n$ for a GP function $\f(\mathbf{x}_n) \in \mathbb{R}^M$, under the setup of intrinsic co-regionalization model (ICM) \cite{alvarez2012kernels}, we have
\begin{align}
    \f(\mathbf{x}_n) = \sum_{i = 1}^S \mathbf{b}_i u^i(\mathbf{x}_n)
\end{align}
where $u^1(\mathbf{x}), \dots, u^S(\mathbf{x})$ are i.i.d. variables following $\mathcal{GP}(0, \mathcal{K}(\mathbf{x,x}'))$ and $\mathbf{b}_i \in \mathbb{R}^M$ for all $i$. This leads to a model whose covariance is
\begin{align}
    \Cov(\f(\mathbf{x}_n), \f(\mathbf{x}_m)) &= \sum_{i=1}^S \sum_{j=1}^S \mathbf{b}_i \mathbf{b}_j^\top \E(u^i(\mathbf{x}_n)u^j(\mathbf{x}_m)) \\
    &= \sum_{i=1}^S \mathbf{b}_i \mathbf{b}_i^\top \E(u^i(\mathbf{x}_n)u^i(\mathbf{x}_m)) \\
    &= \mathcal{K}(\mathbf{x}_n, \mathbf{x}_m) \sum_{i=1}^S \mathbf{b}_i \mathbf{b}_i^\top.
\end{align}
Denoting $\B = (\mathbf{b}_1, \dots, \mathbf{b}_S)$, we can see that $\B\B^\top = \sum_{i=1}^S \mathbf{b}_i \mathbf{b}_i^\top$, thus the covariance can be written as $\Cov(\f(\mathbf{x}_n),\f(\mathbf{x}_m)) = \mathcal{K}(\mathbf{x}_n, \mathbf{x}_m) \B\B^\top$. When we have $N$ input-output data pairs, the full covariance of $\tilde{\f} = \text{vec}(\f(\mathbf{x}_1), \dots, \f(\mathbf{x}_N))$ will follow the separable form $\Cov(\tilde{\f}) = \mathbf{K} \otimes \B\B^\top$. Since a kernel on graphs is usually a square matrix, our graph GP model is equivalent to ICM if $S = M$ and the vectors $\mathbf{b}_i$ combine into a matrix that takes the general form of Eq. (\ref{kernel_on_graph}).

As an additional note, the covariance we derive is dependent on the manner in which $\f(\mathbf{x}_1), \dots, \f(\mathbf{x}_N)$ are stacked into a single vector. If we take $\tilde{\f} = \text{vec}((\f(\mathbf{x}_1), \dots, \f(\mathbf{x}_N))^\top)$ instead, we will get the covariance $\B\B^\top \otimes \mathbf{K}$. These are simply different ways to represent the prior covariance, and $\B\B^\top$ and $\mathbf{K}$ still operate on the output and input space respectively.

\section{Optimizing GP Log-Marginal Likelihood}
\label{sec:opt}

The polynomial coefficients $\beta_i$ in the kernel on graphs are found by maximizing the log-marginal likelihood on a training set using gradient optimization. Let $\boldsymbol{\beta} = (\beta_0, \dots, \beta_P)^\top$, and let $\Omega$ contain $\boldsymbol{\beta}$ and other hyperparameters, the GP log-marginal likelihood is
\begin{align}
    l(\Omega) &= \log \p(\tilde{\mathbf{y}} | \Omega) \\
    &= -\frac{1}{2}\log |\boldsymbol{\Sigma}_\Omega| - \frac{1}{2} \tilde{\mathbf{y}}^\top \boldsymbol{\Sigma}_\Omega^{-1} \tilde{\mathbf{y}} - \frac{NM}{2} \log(2\pi).
    \label{likelihood}
\end{align}
As described in Eq. (\ref{filtering}), the term $\B = g(\mathbf{L}_S)$ also acts as a filter on the GP prior to incorporate information from the graph structure. In order for $\B$ to be a valid filter and a kernel on the graph, we need to constrain $\B$ to be positive semi-definite (PSD); in other words, we need to have $g(\lambda) \geq 0$ for all eigenvalues \cite{shuman2012emerging,alvarez2012kernels}. Just optimizing $\boldsymbol{\beta}$ alone in an unconstrained fashion will not guarantee this, thus we utilize Lagrange multipliers to combine constraints with our main objective function.

Assuming all other hyperparameters are fixed, our constrained optimization problem for finding the optimal kernel on graphs is the following
\begin{align}
\begin{aligned}
    \min_\beta & -l(\boldsymbol{\beta})\\
    \textmd{subject to} \; & - \mathbf{B}_v\boldsymbol{\beta} \leq 0,
\end{aligned} \label{optimprob}
\end{align}
where we express the log-marginal likelihood $l$ as a function of $\boldsymbol{\beta}$ and $\mathbf{B}_v \in \mathbb{R}^{M \times (P+1)}$ is the Vandermonde matrix of eigenvalues of the Laplacian with the following form
\begin{align}
    \mathbf{B}_v = \begin{pmatrix}
    1 & \lambda_1 & \lambda_1^2 & \dots & \lambda_1^{P}\\
    1 & \lambda_2 & \lambda_2^2 & \dots & \lambda_2^{P}\\
    \vdots & \vdots & \vdots & \ddots & \vdots\\
    1 & \lambda_M & \lambda_M^2 & \dots & \lambda_M^{P}
    \end{pmatrix}.
\end{align}
It is easy to see that to have $g(\lambda) \geq 0$ for all eigenvalues is equivalent to setting $-\mathbf{B}_v \boldsymbol{\beta} \leq 0$. Hence, our objective function now becomes
\begin{align}
    \mathbb{L}(\boldsymbol{\beta}, \mathcal{L}) = - l(\boldsymbol{\beta}) + \mathcal{L}^\top (- \B_v \boldsymbol{\beta}) = - l(\boldsymbol{\beta}) - \mathcal{L}^\top \B_v \boldsymbol{\beta}
\end{align}
where $\mathcal{L} \in \mathbb{R}^M$ is a vector of Lagrange multipliers. The solution to this problem is guided by the Karush–Kuhn–Tucker (KKT) conditions \cite{kuhn2014nonlinear}, which specifies that $\boldsymbol{\beta}^*$ is the optimal solution to (\ref{optimprob}) if $(\boldsymbol{\beta}^*, \mathcal{L})$ is the solution to $\min_{\boldsymbol{\beta}} \max_{\mathcal{L}\geq 0} \mathbb{L}(\boldsymbol{\beta}, \mathcal{L})$. Due to the non-convexity of the log-likelihood, the Lagrangian is non-convex with respect to both variables and we instead solve for the dual problem
\begin{align}
    \max_{\mathcal{L}\geq 0} \min_{\boldsymbol{\beta}} \mathbb{L}(\boldsymbol{\beta}, \mathcal{L}).
    \label{optimprob2}
\end{align}
as this makes the function concave with respect to $\mathcal{L}$ \cite{gasimov2002augmented,bazaraa1979survey} leading to an easier problem overall.

We find the solution by alternatively updating $\boldsymbol{\beta}$ and $\mathcal{L}$ described in Algorithm \ref{alg1}. Here, $\mathcal{L}$ is replaced with $e^{\mathcal{L}'}$ and we solve for $\mathcal{L}'$ instead to keep the Lagrange multipliers positive during the optimization.

\begin{algorithm}[t]
\caption{Constrained optimization of polynomial coefficients for GP log-marginal likelihood}
\label{alg1}

\begin{algorithmic}[1]
\STATE {\bfseries Input:} Initialization of $\boldsymbol{\beta}$ and $\mathcal{L}'$
% $\mathbb{L}(\beta, e^{\mathcal{L}'}) = - l(\beta) - e^{\mathcal{L'}^\top} B_v \beta$
\STATE Solve for $\min_\beta \mathbb{L}(\boldsymbol{\beta}, e^{\mathcal{L}'})$ using gradient descent:
$\beta_i \rightarrow \beta_i - \gamma_\beta \frac{\partial \mathbb{L}}{\partial \beta_i}(\boldsymbol{\beta}, e^{\mathcal{L}'})$ for $i = 0, \dots, P$ % until convergence.
\STATE Update $\mathcal{L}'$: $\mathcal{L}' \rightarrow \mathcal{L}' + \gamma_\mathcal{L} \frac{\partial \mathbb{L}}{\partial \mathcal{L}'}(\boldsymbol{\beta}, e^{\mathcal{L}'})$
\STATE Repeat 2 and 3 until $\mathbb{L}$ converges % in both $\boldsymbol{\beta}$ and $\mathcal{L}'$
\STATE {\bfseries Output:} $\boldsymbol{\beta}$
\end{algorithmic}

\end{algorithm}

Due to the non-convexity of (\ref{optimprob2}), Algorithm \ref{alg1} may only find a local optimum depending on the initialization. A simple strategy to obtain a sensible initialization is to optimize for the log-marginal likelihood (without the constraint on $\boldsymbol{\beta}$) using gradient ascent, with initializations chosen from a small set of values that lead to the highest log-marginal likelihood. The solution to this unconstrained optimization is then used as the initialization for Algorithm \ref{alg1}. The algorithm is much more stable with respect to the initialization of the Lagrange multipliers, and using either a fixed or random initialization works well in practice.%, with the only rule of thumb being to keep the initializations small so the Lagrange multipliers terms do not dominate the log-likelihood.

\subsection{Scalability}

By exploiting the Kronecker product structure of the covariance matrix, inversion of eq. (\ref{input_output_space}) needed for Algorithm \ref{alg1} and GP inference can be reduced to a runtime of $\mathcal{O}(N^3 + M^3)$ and thus avoiding the expensive $\mathcal{O}(N^3M^3)$. We manipulate the matrix in a similar fashion to \cite{pu2020kernel}, with the derivation tailored to our model left to supplementary. Potential further reduction through sparse variational inference will be left as future work.

\section{Related Work}

Learning on graph-structured data has been studied from both machine learning and signal processing perspectives such as \cite{shuman2012emerging,ortega2018graph,bronstein2017geometric,wu2019comprehensive}. Our model is unique in that it makes use of tools from both fields to achieve interpretations of filtering and kernel learning in the graph spectral domain.

Laplacian-based functions in graph signal processing such as graph filters have been applied to data with certain smoothness assumptions, thus transforming data into one of low-, band- or high-pass profiles \cite{shuman2012emerging,ortega2018graph}. In contrast, our algorithm learns the filter based on the data to exempt the need for choosing the filter profile a priori. This extends the non-probabilistic approach in \cite{thanou2013parametric} with the added benefit of producing a measure of uncertainty. %On the other hand, some learning frameworks in the graph spectral domain such as \cite{bruna2013spectral} requires eigen-decomposition of the Laplacian which is of $O(M^3)$, while our model avoids this problem as the kernel spectrum is specified without explicitly going into the graph spectral domain. 
%Graph neural networks that are comparable to our model are spectral based convolutional models such as \cite{defferrard2016convolutional,kipf2016semi,Bruna2014SpectralNA} %avoid this issue but 
%these typically require a large amount of training data; in comparison, our model is data-efficient and only requires a small number of training signals as shown in the experimental results.

The kernel defined on the outputs also bears similarity to spectral designs of graph neural networks (GNNs) such as \cite{defferrard2016convolutional,kipf2016semi, Bruna2014SpectralNA}. The work in \cite{Bruna2014SpectralNA} proposes to learn a free-form graph filter, which does not guarantee its spatial localization. The models in \cite{defferrard2016convolutional} and \cite{kipf2016semi} do offer such localization property; however, they typically require a large amount of training data. In comparison, our model is more data-efficient and only requires a small number of training signals as demonstrated by the experimental results.

In previous works relating to GP on graphs, our model resembles that of \cite{venkitaraman2018gaussian}, but with the distinction that the kernel on the output space is learnt instead of a chosen low-pass filter. Algorithm \ref{alg1} also demonstrates that spectral kernel learning, which has been applied to learning continuous kernels \cite{Wilson2013GaussianPK}, is also possible on a discrete graph space. Other models in this field are predominantly applied to scalar outputs problems. The way the graph is utilized follows a similar framework to graph neural network models such as \cite{kipf2016semi}, with one representative approach being local averaging of neighbourhood information for node level classification \cite{ng2018bayesian}. More complicated aggregation functions have since been applied as a linear function to the GP covariance in \cite{opolka2020graph, chengdynamic, liuuncertainty}. Although these models may be extended to a vector output, they generally involve averaging or smoothing of the data, and the resulting effect is similar to a low-pass filter on the graph. Hence, these models are likely to perform less well on data that are not customarily smooth. Our model overcomes this issue through spectral learning of the kernel on graphs to adapt to the data more effectively. Finally, the convolutional patch technique in \cite{van2017convolutional} has also been extended to graph data \cite{walker2019graph}. This method can be viewed as an extension to the approach in \cite{ng2018bayesian}, but it is still based on pre-defined kernel functions in the graph domain.

\begin{figure*}[t]
\centering
\begin{subfigure}[b]{0.30\linewidth}
\includegraphics[width=\linewidth, height = 3.8cm]{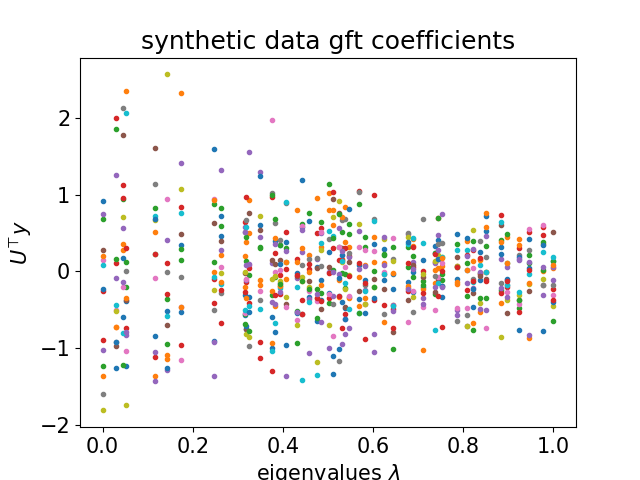}\caption{}\label{low_pass1}
\end{subfigure}
\begin{subfigure}[b]{0.30\linewidth}
\includegraphics[width=\linewidth, height = 3.8cm]{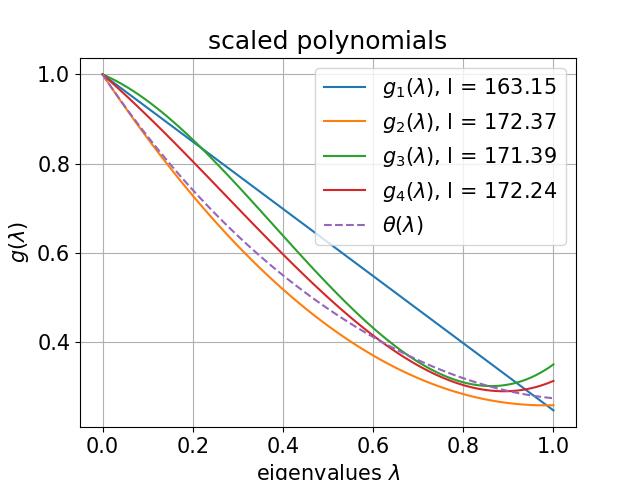}
\caption{}\label{low_pass2}
\end{subfigure}
\begin{subfigure}[b]{0.30\linewidth}
\includegraphics[width=\linewidth, height = 3.8cm]{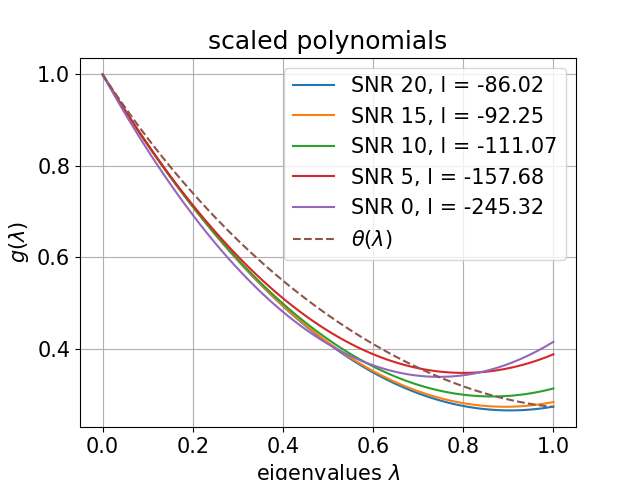}
\caption{}\label{noise1}
\end{subfigure}
\begin{subfigure}[b]{0.30\linewidth}
\includegraphics[width=\linewidth, height = 3.8cm]{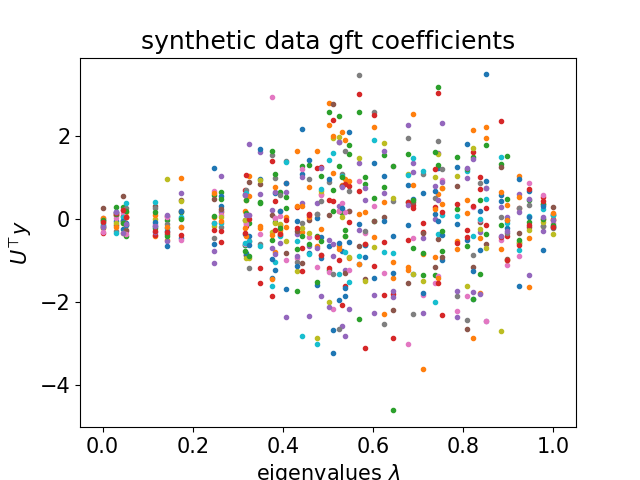}
\caption{}\label{band_pass1}
\end{subfigure}
\begin{subfigure}[b]{0.30\linewidth}
\includegraphics[width=\linewidth, height = 3.8cm]{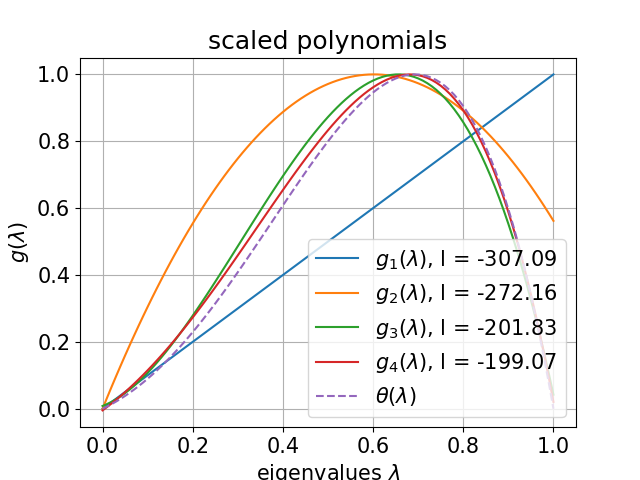}
\caption{}\label{band_pass2}
\end{subfigure}
\begin{subfigure}[b]{0.30\linewidth}
\includegraphics[width=\linewidth, height = 3.8cm]{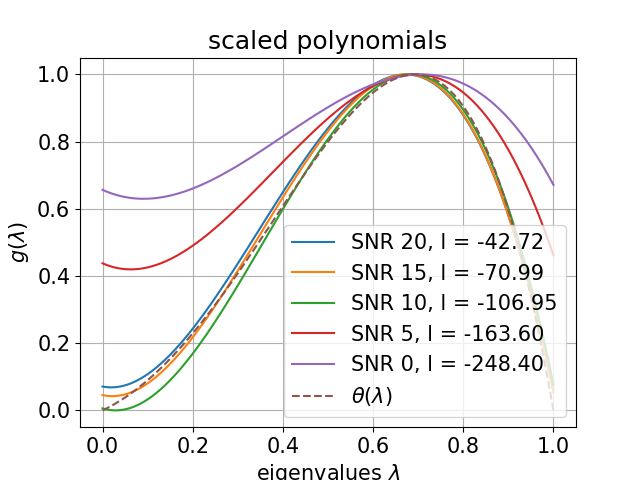}
\caption{}\label{noise2}
\end{subfigure}
\vspace{-0.2cm}
\caption{Spectral kernel learning on synthetic data, where we show the synthetic graph Fourier coefficients, and the scaled polynomials learnt with their log-marginal likelihoods, for data with low-pass spectrum (a)-(c) and band-pass spectrum (d)-(f). Ground truth polynomials are $\theta = (1., -1.5, 1.5^2/2., -1.5^3/6, 1.5^4/24)$ for the low-pass (first 5 terms of $e^{-1.5}$), and $\theta = (0, 1, 4, 1, -6)$ for the band-pass.} % Separately, learning is also done with varying noise SNR using a degree 2 polynomial for low-pass and degree 3 polynomial for band-pass in (c) and (f).} 
\label{low_pass}
\end{figure*}

\section{Experiments} \label{experiments}

In this section, we first present results on synthetic experiments to demonstrate our algorithm's ability to recover ground truth filter shapes. We then apply our method to several real-world datasets that exhibit different spectral characteristics to show the adaptability of our model.

In all experiments, the GP prior will be in the form of Eq. (\ref{multigau_local}) and we consider baseline GP models from \cite{venkitaraman2018gaussian, ng2018bayesian}, and kernels on graphs defined in \cite{smola2003kernels} in Eq 17-20:
\begin{itemize}
    \item Standard GP $\B = \mathbf{I}$
    \item Global filtering $\B = (\mathbf{I} + \alpha \mathbf{L})^{-1}$ \cite{venkitaraman2018gaussian}
    \item Local averaging $\B = (\mathbf{I} + \alpha \mathbf{D})^{-1}(\mathbf{I} + \alpha \mathbf{A})$ \cite{ng2018bayesian} where we also added a weighting parameter $\alpha$.
    \item Graph Laplacian regularization $\B\B^\top = \mathbf{L}^\dagger$ (pseudo-inverse of the Laplacian) \cite{alvarez2012kernels}
    \item Regularized Laplacian $\mathbf{BB}^\top = (\mathbf{I}+ \alpha \tilde{\mathbf{L}})^{-1}$ \cite{smola2003kernels}
    \item Diffusion $\mathbf{BB}^\top = \exp\{(-\alpha/2) \tilde{\mathbf{L}}\}$ \cite{smola2003kernels}
    \item $p$-step random walk $\mathbf{BB}^\top = (\alpha\mathbf{I} - \tilde{\mathbf{L}})^p$ \cite{smola2003kernels}
    \item Cosine $\mathbf{BB}^\top = \cos (\tilde{\mathbf{L}}\pi/4)$ \cite{smola2003kernels}
\end{itemize}
The input kernel will be squared exponential $\mathbf{K}_{ij} = \sigma^2_w \exp \{-\frac{1}{2l}||\mathbf{x}_i - \mathbf{x}_j||^2_2\}$ applied to inputs $\mathbf{x}_i$ and $\mathbf{x}_j$, giving a total set of hyperparameters to be $\Omega = \{\boldsymbol{\beta}, \alpha, l, \sigma_w^2, \sigma_\epsilon^2\}$. All hyperparameters in the baselines are found by maximizing the log-marginal likelihoods by gradient ascent, with the exception that we optimize $\beta$ using Algorithm 1, with the initialization described at the end of Section \ref{sec:opt} (further details are presented in Supplementary Material). The predictive performance will be evaluated by the posterior log-likelihoods $\log \p(\mathbf{y}_* | \boldsymbol{\mu}_*, \boldsymbol{\Sigma}_*)$ for test signals $\mathbf{y}_*$, with GP posterior mean $\boldsymbol{\mu}_*$ and covariance $\boldsymbol{\Sigma}_*$.

We would also like to investigate the effect of the constraint on learning performance. To this end, we include a baseline where we only solve the problem of (18) without the constraint. In the real-world experiments, we include this baseline using a degree-3 polynomial, where the resulting spectral functions are generally not always valid graph filtering functions.

\begin{figure*}[t]
\centering
\begin{subfigure}[b]{0.30\linewidth}
\includegraphics[width=\linewidth, height = 3.8cm]{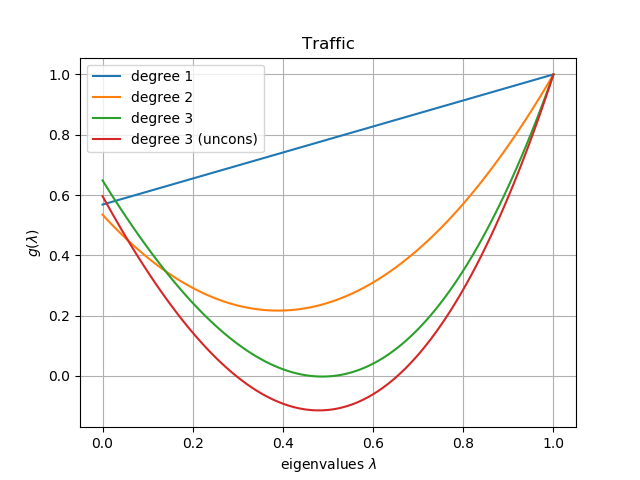}
\caption{}\label{filter_traffic}
\end{subfigure}
\begin{subfigure}[b]{0.30\linewidth}
\includegraphics[width=\linewidth, height = 3.8cm]{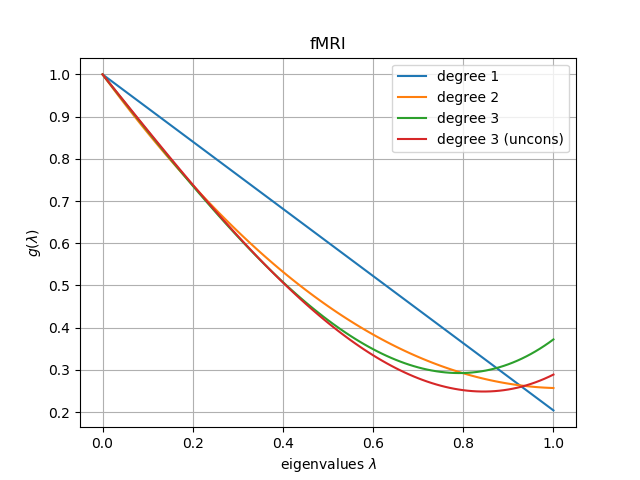}
\caption{}\label{filter_fmri}
\end{subfigure}
\begin{subfigure}[b]{0.30\linewidth}
\includegraphics[width=\linewidth, height = 3.8cm]{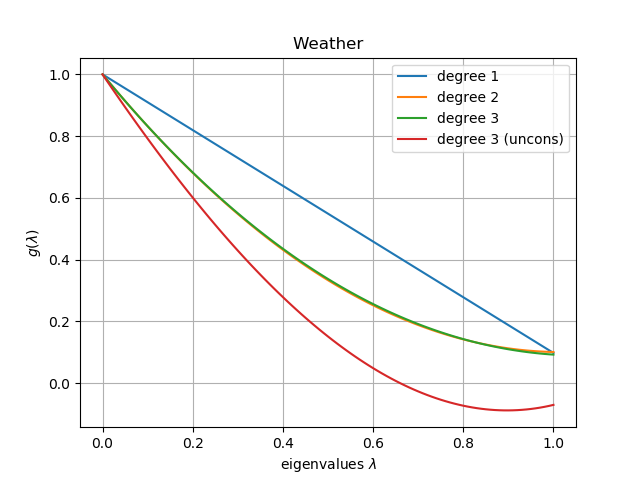}
\caption{}\label{filter_weather}
\end{subfigure}
\vspace{-0.2cm}
\caption{spectra of kernel on graphs learnt using degree 1, 2 and 3 polynomials on the 3 real-world datasets. Each plot is the spectrum learnt on the larger training set from Table \ref{traffic_fmri}.} \label{filter_data}
\end{figure*}

\begin{table*}[t]
  \caption{Test log-likelihoods (higher the better) and standard error in bracket. Results are averaged from 10 test sets, of size 10 for traffic data, size 25 for fMRI and size 6 for weather.} \label{traffic_fmri}
  \centering
  \tabcolsep=0.03cm
  %\scalebox{0.8}{
  \begin{tabular}{lcccccc}
    \hline
\textbf{MODEL} & \textbf{TRAFFIC} & \textbf{TRAFFIC} & \textbf{fMRI} & \textbf{fMRI}  & \textbf{WEATHER} & \textbf{WEATHER}\\
(Training) & (10 signals) & (20 signals) & (21 signals) & (42 signals) & (15 signals) & (30 signals) \\
\hline \\
Degree 1 Polynomial & -324.79 (2.77) & -323.06 (2.89) & 42.36 (4.59) & 45.24 (3.74) & -5.10 (4.16) & -0.43 (3.45) \\
%\hline
Degree 2 Polynomial & \textbf{-322.78 (3.39)} & -321.90 (3.26) & \textbf{42.86 (4.48)} & 45.24 (3.74) & -0.50 (3.37) & \textbf{2.78 (3.11)} \\
%\hline
Degree 3 Polynomial & -324.36 (3.43) & \textbf{-321.76 (3.20)} & 42.77 (4.50) & 46.00 (3.56) & \textbf{-0.32 (3.39)} & 2.51 (3.11) \\
%\hline
Degree 3 (unconstrained) & -324.93 (2.69) & -323.09 (3.55) & 42.70 (4.31) & \textbf{46.32 (3.42)} & -1.58 (3.51) & 1.23 (4.17) \\
\hline
Standard GP & -331.48 (3.03) & -331.80 (3.22) & 39.10 (4.97) & 44.21 (3.79) & -21.73 (6.01) & -20.44 (4.16) \\
%\hline
Laplacian \cite{alvarez2012kernels} & -330.14 (1.22) & -330.38 (1.28) & - & - & -54.29 (6.50) & -54.28 (6.45)\\
%\hline
Global Filtering \cite{venkitaraman2018gaussian} & -330.87 (1.05) & -331.37 (1.01) & 42.50 (4.58) & 45.17 (3.76) & -5.88 (3.09) & -1.28 (2.94) \\
%\hline
Local Averaging \cite{ng2018bayesian} & -330.13 (1.22) & -332.66 (0.91) & 40.62 (4.92) & 43.57 (3.95) & -9.76 (3.76) & -3.31 (3.47) \\

Regularized Laplacian \cite{smola2003kernels} & -325.35 (2.57)  & -324.39 (3.17) & 38.85 (5.63) & 44.20 (3.46) & -2.00 (3.33) & 1.22 (3.41)\\

Diffusion \cite{smola2003kernels} & -328.21 (2.67) & -328.20 (2.86) & 40.92 (4.26) & 41.97 (3.76) & -3.33 (3.17) & 2.08 (2.86)\\

1-Step Random Walk \cite{smola2003kernels} & -324.98 (2.29) & -322.24 (3.09) & 41.22 (4.34) & 40.79 (4.13) & -21.52 (5.70) & -20.78 (4.13)\\

3-Step Random Walk \cite{smola2003kernels} & -324.75 (2.35) & -323.67 (2.75) & 41.54 (4.79) & 45.96 (3.91) & -4.10 (4.56) & 0.97 (3.98)\\

Cosine \cite{smola2003kernels} & -323.48 (2.54) & -322.23 (2.74) & 33.95 (6.29) & 42.17 (3.80) & -17.93 (7.38) & -6.71 (7.07)\\
    \hline
  \end{tabular}%}
\end{table*}

\subsection{Synthetic Signals} \label{synthetic}

For the first experiment we use synthetic signals which are generated following Eq. (\ref{filtering}) using a $\mathbf{B}$ with a known polynomial chosen beforehand. The aim is to demonstrate that our model can recover the polynomial shapes of the ground truth filters through optimizing the GP log-marginal likelihood.

We set the underlying graph to be a 30 nodes Sensor graph from the \url{pygsp} library \cite{defferrard2017pygsp}. The Sensor graph has an even spread of eigenvalues which helps the visualization of the polynomial (further results on other random graph models are presented in Supplementary Material). Signals are first sampled independently as $\mathbf{y}_1', \mathbf{y}_2', \dots \sim \N(0,\mathbf{I})$. Using the scaled graph Laplacian $\mathbf{L}_S$, we denote the ground truth filter as $\theta(\mathbf{L}_S)$ with coefficients $(\theta_0, \dots, \theta_Q)$. Each synthetic signal is then set as $\mathbf{y}_i = \theta(\mathbf{L}_S)\mathbf{y}_i'$ and we corrupt it with noise at a signal-to-noise ratio (SNR) of 10 dB. As the signals are sampled independently, the kernel function is $\B\B^\top \otimes \sigma^2_w\mathbf{I} + \sigma^2_\epsilon\mathbf{I}$ where $\sigma^2_w$ is set to signal variance. 
% The noise variance $\sigma_\epsilon^2$ is fixed as the solution of an initial optimization run, so each time we only learn the coefficients. 
We denote the polynomials learnt from our algorithm as $g_d$ for degree $d$ which has $d+1$ coefficients. If the $g_d(\lambda)$ goes above 1 for any $\lambda \in [0,1]$, we can scale it down as $g_d'(\lambda) = \frac{1}{c} g_d(\lambda)$ for $c = \max_{x\in[0,1]} g_d(x)$. The resulting $g_d'$ will be in the range $[0,1]$ making it easier to compare different filters, and the $c$ term can be absorbed into the variance of the full kernel function, alleviating the need to optimize for $\sigma^2_w$ in $\mathbf{K}$.

In Fig. \ref{low_pass}, we show the results from learning on synthetic data with low- and band-pass spectrum (a high-pass spectrum will simply have the reversed shape of the low-pass so we will not present here due to the similarity). In Fig. \ref{low_pass1} and Fig. \ref{band_pass1} we plot the graph Fourier coefficients $\mathbf{U^\top y}$ of the generated signals, with each colour corresponding to one signal. % these values can be both positive and negative so we see the shape of $\theta(\lambda)$ reflected about $y = 0$. 
The learnt polynomials with different degrees can be found in Fig. \ref{low_pass2} and \ref{band_pass2} along with the ground truth polynomial $\theta(\cdot)$. Visually we can see that using a polynomial with $d = 2$ and $3$ respectively capture the ground truths of low- and high-pass filters well enough that higher degree no longer offers clear improvement. This is also evident in the log-marginal likelihoods, where we see only little improvement for $d > 2$ for low-pass and $d > 3$ for band-pass spectra. 

We next study the effect of noise on learning the spectrum, using a degree 2 polynomial for low-pass and degree 3 for band-pass. % we now update both $\boldsymbol{\beta}$ and $\sigma_\epsilon^2$ in the optimization as the amount of noise is different each time. 
Fig. \ref{noise1} and \ref{noise2} show the spectrum learnt for various SNRs, where we can see visually that our model recovers the true spectrum well for SNR 10 dB or higher. As expected, the corresponding marginals steadily decrease as SNR decreases when data becomes noisier.

\subsection{Traffic Dataset}

In the first real-world experiment we consider the daily traffic bottleneck stations in San Francisco \cite{choe2002freeway, thanou2013parametric}. Stations corresponding to nodes are connected in the graph if Euclidean distances are less than a threshold of 13 kilometers with inverse distance as edge weights. The signal is the average time (in minutes) that each bottleneck is active on a specific day. The graph consist of 75 nodes, where we use the data on the first 15 nodes as input $\mathbf{x}_n$, and predict on the remaining 60 nodes giving us $\mathbf{y}_n \in \mathbb{R}^{60}$. We find the hyperparameters on the training set followed by conditioning to compute the posterior. The test signals are split 10-fold, with posterior log-likelihoods computed on each fold to provide an average test log-likelihood and standard error. The graph signals exhibit large Fourier coefficients for eigenvalues near 0 and 1, making the baseline models less suitable. The results are found in the first two columns of Table \ref{traffic_fmri} for a training set of 10 and 20 signals, and the shapes of the kernel spectrum are presented in Fig. \ref{filter_traffic}. We notice that the performance of the constrained optimization is better than that of the unconstrained version which was not fully non-negative. This suggests that enforcing positivity may have helped the model generalize to testing data.

\subsection{fMRI Dataset}

We next consider data from functional magnetic resonance imaging (fMRI) where an existing graph of 4465 nodes corresponds to different voxels of the cerebellum region in the brain (we refer to \cite{venkitaraman2018gaussian,behjat2016signal} for more details on graph construction and signal extraction). A graph signal is the blood-oxygen-level-dependent (BOLD) signal observed on the voxels. We use the graph of the first 50 nodes, taking the readings on the first 10 nodes as $\mathbf{x}_n$ to predict the outcome signals $\mathbf{y}_n$ on the remaining 40 nodes ($\mathbf{y}_n \in \mathbb{R}^{40}$). The dataset contains 292 signals for which we train on a sample of up to 42 signals to learn the hyperparameters. We then compute the posterior to predict the remaining 250 signals, which are split 10-fold to provide a mean test log-likelihood and standard error. This dataset follows a low-pass spectrum that suits kernels in the baseline models, in particular global filtering \cite{venkitaraman2018gaussian}, thus our algorithm is expected to produce similar performances. The results can be found in the middle two columns of Table \ref{traffic_fmri} and the learnt polynomials in Fig. \ref{filter_fmri}, we see the posterior log-likelihoods of our models for degree 2 and 3 are still producing the best performances, but the unconstrained degree 3 polynomial was able to outperform all models for 42 training signals. This is likely because the unconstrained spectrum is already positive, thus Algorithm \ref{alg1} made little change to the spectral function and the performance did not always improve.

\subsection{Weather Dataset}

The last dataset is the temperature measurement in 45 cities in Sweden available from the SMHI \cite{swedish}. The data also follows a low-pass spectrum making the baselines from \cite{ng2018bayesian} and \cite{venkitaraman2018gaussian} suitable models. Using the cities' longitude and latitude, we construct a $k$-nearest neighbour graph for $k = 10$ using \url{pygsp} \cite{defferrard2017pygsp}. We perform the task of next-day prediction, where given $\mathbf{x}_n \in \mathbb{R}^{45}$ as the temperature signal at day $n$, we aim to predict $\mathbf{y}_n \in \mathbb{R}^{45}$ as the temperature signal on day $n+1$. We randomly sample 30 signal pairs $(\mathbf{x}_n, \mathbf{y}_n)$ for hyperparameter learning, and predict the signals on the remaining 60 days. The results are in the final two columns of Table \ref{traffic_fmri} and filter shapes in Fig. \ref{filter_weather}, we see degree 2 and 3 polynomials are both performing better than other models, while the unconstrained polynomial is still slightly lower much like the traffic results.

\section{Conclusion}

We have developed a novel GP-based method for graph-structured data to capture the inter-dependencies between observations on a graph. The kernel on graphs adapts to the characteristics of the data by using a bespoke learning algorithm that also provides a better interpretability of the model from a graph filtering perspective. Our model has produced superior performances on highly non-smooth data while results were competitive with the baselines on data that are generally smoother. Promising future directions include the extension of the model for application in classification and improvement in scalability of the model.

%\subsubsection*{References}

\bibliographystyle{unsrt}
%\bibliography{bib/References.bib}
\bibliography{main.bbl}

\newpage
\onecolumn
\setcounter{section}{0}
\setcounter{equation}{0}

\aistatstitle{Gaussian Processes on Graphs via Spectral Kernel Learning:\\ Supplementary Materials}

\section{Scalability}

MOGP inference requires inverting covariance matrix $\mathbf{\Sigma} = \mathbf{BB}^\top \otimes \mathbf{K} + \sigma^2_\epsilon \mathbf{I}$ which is of $\mathcal{O}(N^3M^3)$, but by exploiting the structure of the Kronecker product we can reduce the computation to $\mathcal{O}(N^3 + M^3)$.

We re-write the covariance matrix as follows
\begin{align}
    \mathbf{\Sigma} &= \mathbf{BB}^\top \otimes \mathbf{K} + \sigma^2_\epsilon \mathbf{I}\\
    &= (\mathbf{I} \otimes \mathbf{K})(\mathbf{BB}^\top \otimes \mathbf{I}) + \sigma^2_\epsilon (\mathbf{BB}^\top \otimes \mathbf{I})^{-1} \mathbf{BB}^\top \otimes \mathbf{I}\\
    &= \big[\mathbf{I} \otimes \mathbf{K} + \sigma^2_\epsilon ((\mathbf{BB}^\top)^{-1} \otimes \mathbf{I})\big] \mathbf{BB}^\top \otimes \mathbf{I}\\
    &= \big[\sigma^2_\epsilon (\mathbf{BB}^\top)^{-1} \oplus \mathbf{K}\big] \mathbf{BB}^\top \otimes \mathbf{I}
\end{align}
for Kronecker sum $\oplus$. Next, take the eigen-decomposition $\mathbf{K} = \mathbf{U}_K\mathbf{\Lambda}_K\mathbf{U}_K^\top$ and $\mathbf{BB}^\top = \mathbf{U}_B\mathbf{\Lambda}_B\mathbf{U}_B^\top$, the above equation becomes
\begin{align}
    \mathbf{\Sigma} &= \big[\sigma^2_\epsilon \mathbf{U}_B\mathbf{\Lambda}_B^{-1}\mathbf{U}_B^\top \oplus \mathbf{U}_K\mathbf{\Lambda}_K\mathbf{U}_K^\top \big]\mathbf{U}_B\mathbf{\Lambda}_B\mathbf{U}_B^\top \otimes \mathbf{I}\\
    &= \sigma^2_\epsilon(\mathbf{U}_B \otimes \mathbf{U}_K) (\mathbf{\Lambda}^{-1}_B \oplus \mathbf{\Lambda}_K) (\mathbf{U}_B^\top \otimes \mathbf{U}_K^\top) (\mathbf{U}_B\mathbf{\Lambda}_B\mathbf{U}_B^\top \otimes \mathbf{I}).
\end{align}
Each bracket can then be individually inverted by utilizing the orthogonality of the eigen matrices and the full matrix inverse becomes
\begin{align}
    \mathbf{\Sigma}^{-1} = \frac{1}{\sigma^2_\epsilon} (\mathbf{U}_B\mathbf{\Lambda}_B^{-1}\mathbf{U}_B^\top \otimes \mathbf{I}) (\mathbf{U}_B \otimes \mathbf{U}_K) (\mathbf{\Lambda}_B \oplus \mathbf{\Lambda}^{-1}_K) (\mathbf{U}_B^\top \otimes \mathbf{U}_K^\top).
\end{align}
Computational complexity is therefore dominated by the two eigen-decomposition matrices of size $N\times N$ and $M\times M$ giving an overall cost of $\mathcal{O}(N^3 + M^3)$.

\section{Initialization Strategy}

Due to the highly non-convex structure of the GP log-marginal likelihood, optimizing hyperparameters is heavily reliant on the initializations. Here, we propose a procedure of steps to get the best and most stable solution for Algorithm 1. We are aware that there may be more generalizable initialization strategies; what is presented here is one that we found worked well for our problem.

Based on a training set $\{\mathbf{y}_1,\dots,\mathbf{y}_N\}$, the set of hyperparameters to learn is $\Omega = \{\boldsymbol{\beta}, l, \sigma_w^2, \sigma_\epsilon^2\}$, from which we initialize
\begin{align}
    l &= \textmd{Mean}(\{||\mathbf{y}_1||_2^2, \dots, ||\mathbf{y}_N||_2^2\})\\ \sigma^2_w &= \Var(\{\mathbf{y}_1, \dots, \mathbf{y}_N\}).
\end{align}
We set the other parameters by trying a small range of values, using the combination that leads to the highest log-marginal likelihood as the initialization. Our procedure is as follows:
\begin{enumerate}
    \item Find the optimal $\boldsymbol{\beta}$ and $\sigma_\epsilon^2$ that maximizes the log-marginal likelihood by a grid search.
    \item Use the best combinations from grid search as initializations (along with initial $l$ and $\sigma_w^2$) for the unconstrained problem - maximizing the log-marginal likelihood with respect to $\{\boldsymbol{\beta}, l, \sigma_\epsilon^2\}$ ($\sigma_w^2$ is indirectly optimized through $\boldsymbol{\beta}$ as explained in Section 6.1).
    \item Use the solution found in step 2 as the initializations to Algorithm 1 and solve for $\boldsymbol{\beta}$, while keeping all other hyperparameters constant.
\end{enumerate}

\newpage

In the grid search we use $\sigma_\epsilon^2 \in \{\frac{1}{10}\sigma^2_w, \frac{1}{5}\sigma^2_w \}$, while for each elements $\beta_i$ we use $\beta_i \in \{-5,-4,\dots,5\}$ for low-pass, and $\beta_i \in \{-10,-8,\dots,10\}$ for more complicated spectrums such as band-pass and that of the traffic data.

As a final note, we follow a general rule for selecting the learning rate for each hyperparameter ($\gamma_\beta, \gamma_l$, etc) as: choosing the largest $r \in \mathbb{Z}$ such that $\gamma_p = 10^r$ for hyperparameter $p$, that leads to a consistent increase/decrease in the objective function. This will require some tuning from the user beforehand in order to ensure the algorithm converges in a reasonable time.

\section{Synthetic Experiments using Barabási–Albert Random Graph}

\begin{figure}[t]
\centering
\begin{subfigure}[b]{0.38\linewidth}
\includegraphics[width=\linewidth, height = 3.7cm]{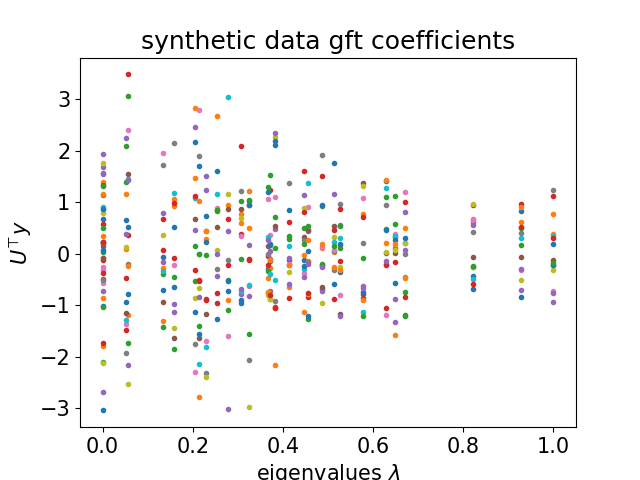}\caption{}\label{ba_low_pass1}
\end{subfigure}
\begin{subfigure}[b]{0.38\linewidth}
\includegraphics[width=\linewidth, height = 3.7cm]{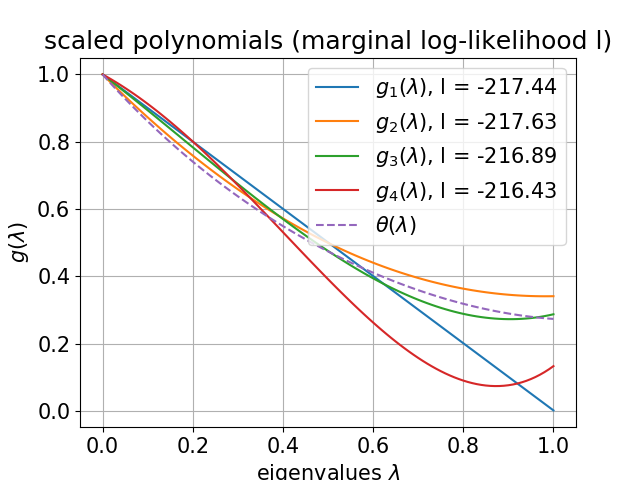}
\caption{}\label{ba_low_pass2}
\end{subfigure}

\begin{subfigure}[b]{0.38\linewidth}
\includegraphics[width=\linewidth, height = 3.7cm]{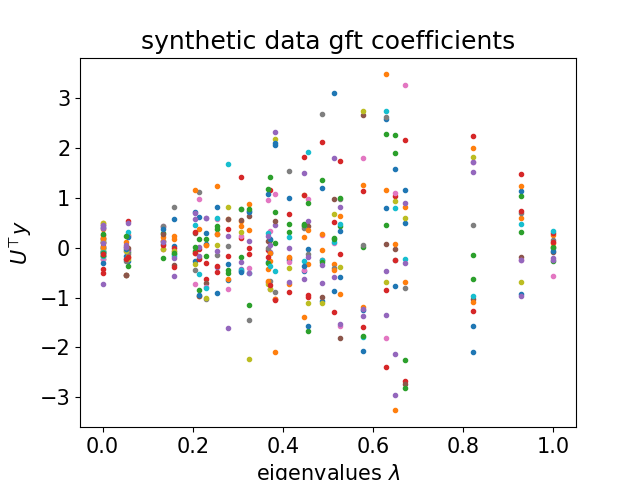}
\caption{}\label{ba_band_pass1}
\end{subfigure}
\begin{subfigure}[b]{0.38\linewidth}
\includegraphics[width=\linewidth, height = 3.7cm]{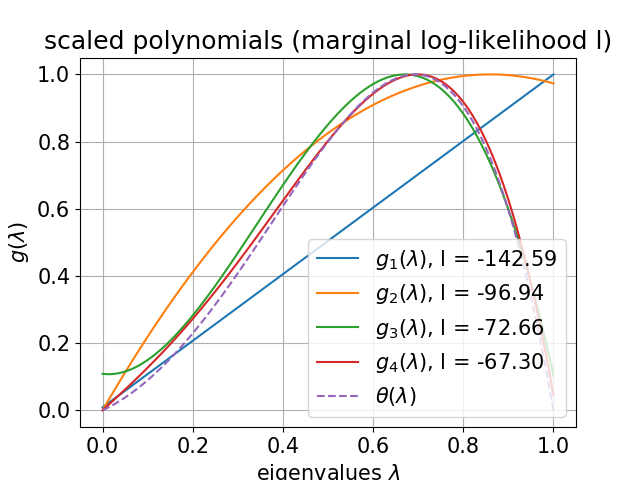}
\caption{}\label{ba_band_pass2}
\end{subfigure}
\vspace{-0.2cm}
\caption{Synthetic experiments on BA graph for low-pass (a)-(b) and band-pass (c)-(d) filters. Ground truth polynomials are $\theta = (1., -1.5, 1.5^2/2., -1.5^3/6, 1.5^4/24)$ for the low-pass (first 5 terms of $e^{-1.5}$), and $\theta = (0, 1, 4, 1, -6)$ for the band-pass filter.}
\label{ba}
\end{figure}

In order to show that our algorithm can generalize to different graphs, we also tried recovering graph filters using on a Barabási–Albert (BA) random graph in place of the Sensor graph described in Section 6.1. The graph contains 30 nodes generated with an initial 10 nodes, and each node added will be randomly connected to 5 existing nodes. Signals are generated using the same form as in the Sensor graph example, as well as using the same full GP kernel function. In Fig. \ref{ba_low_pass1} and \ref{ba_band_pass1} we can visually see the distribution of the eigenvalues are less uniform, while Fig. \ref{ba_low_pass2} and \ref{ba_band_pass2} show the low- and band-pass filter shapes are still recovered relatively well using Algorithm 1.

\section{Synthetic Experiments with Pre-Defined Kernel on Input Space}

In our synthetic experiments in Section 6.1, we use $\mathbf{K} = \mathbf{I}$ due to the signals being sampled independently. This allows us to see the full effect of the kernel on the output space, but cannot be used for inference as predictions from the posterior will be the same as the prior. We next present a more comprehensive synthetic generative model that defines the covariance between signals. This will also act as a sanity check for our algorithm's ability to recover graph filters when paired with different input kernels.

\begin{figure}[t]
\centering
\begin{subfigure}[b]{0.38\linewidth}
\includegraphics[width=\linewidth, height = 3.7cm]{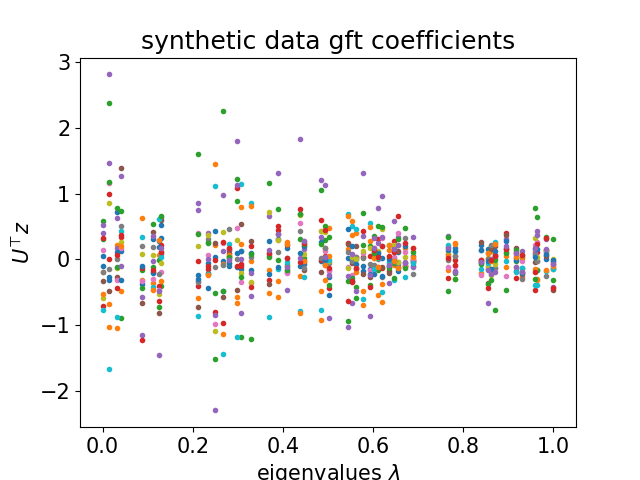}\caption{}\label{low_wishart1}
\end{subfigure}
\begin{subfigure}[b]{0.38\linewidth}
\includegraphics[width=\linewidth, height = 3.7cm]{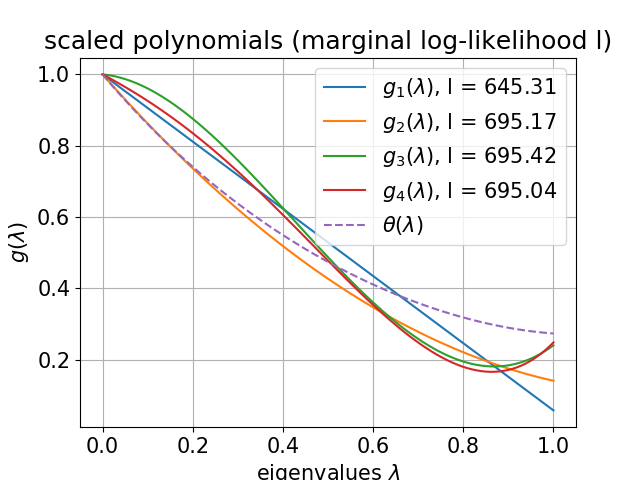}
\caption{}\label{low_wishart2}
\end{subfigure}

\begin{subfigure}[b]{0.38\linewidth}
\includegraphics[width=\linewidth, height = 3.7cm]{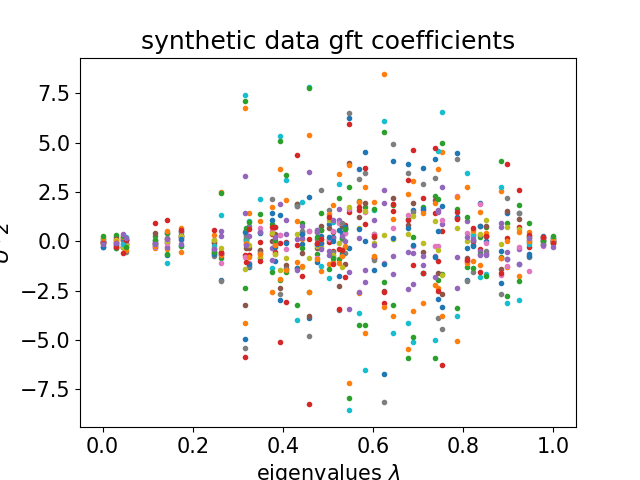}
\caption{}\label{band_wishart1}
\end{subfigure}
\begin{subfigure}[b]{0.38\linewidth}
\includegraphics[width=\linewidth, height = 3.7cm]{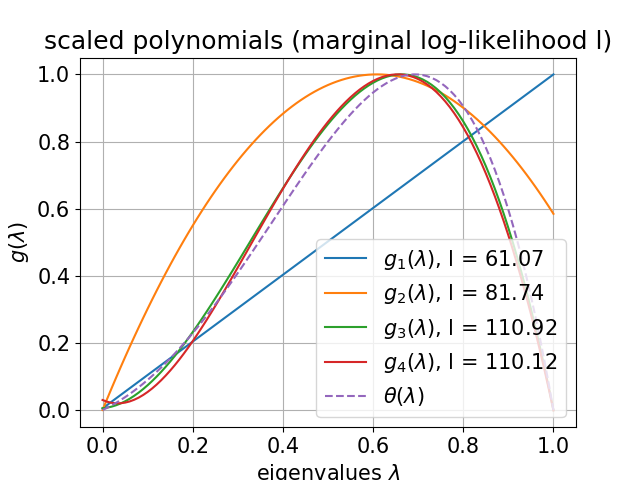}
\caption{}\label{band_wishart2}
\end{subfigure}
\vspace{-0.2cm}
\caption{Synthetic experiments on Sensor graph with an inverse Wishart sample as kernel on input space for low pass (a)-(b) and band pass (c)-(d). Ground truth polynomials are $\theta = (1., -1.5, 1.5^2/2., -1.5^3/6, 1.5^4/24)$ for the low-pass (first 5 terms of $e^{-1.5}$), and $\theta = (0, 1, 4, 1, -6)$ for the band-pass.}
\label{wishart}
\end{figure}

We start by sampling a positive definite matrix $\mathbf{C}$ from inverse Wishart distribution with identity hyperparameter, where $\mathbf{C}$ is of size $N\times N$. We then draw $M$ samples from $\mathcal{N}(0, \mathbf{C})$ to create our data matrix $\Delta$ of size $N\times M$. Each row in $\Delta$ is of dimension $M$ and so can be interpreted as a signal on the graph. Next, let $\mathbf{r}_i$ denote the $i$th row of $\Delta$, we filter this signal by
\begin{align}
    \mathbf{y}_i = \theta(\mathbf{L}_S)\mathbf{r}_i.
\end{align}
The signals $\mathbf{y}_i$ and $\mathbf{y}_j$ of $\Delta$ will have covariance $\mathbf{C}_{ij}\B\B^\top$, and elements of each $\mathbf{y}_i$ will be determined by the filter $\theta(\mathbf{L}_S)$. Hence $\mathbf{C}$ can be used as the covariance matrix on the input space, and the full kernel of the GP becomes
\begin{align}
    \mathbf{C} \otimes \B\B^\top + \sigma_\epsilon^2 \mathbf{I}.
\end{align}
We present the recovered filter shapes in Fig. \ref{wishart} for the two previously used ground truth filter shapes on a Sensor graph. The results are similar as degree 2 and 3 polynomials are able to respectively recover low- and band-pass filters well, both visually and in terms of log-marginal likelihood.

\section{Additional Information for Real-World Experiments}

\begin{figure}[h]
\centering
\begin{subfigure}[b]{0.32\linewidth}
\includegraphics[width=\linewidth, height = 3.8cm]{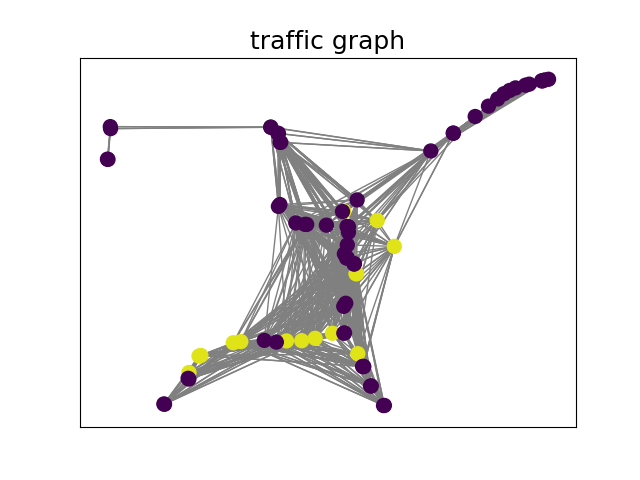}
\caption{}\label{traffic1}
\end{subfigure}
\begin{subfigure}[b]{0.32\linewidth}
\includegraphics[width=\linewidth, height = 3.8cm]{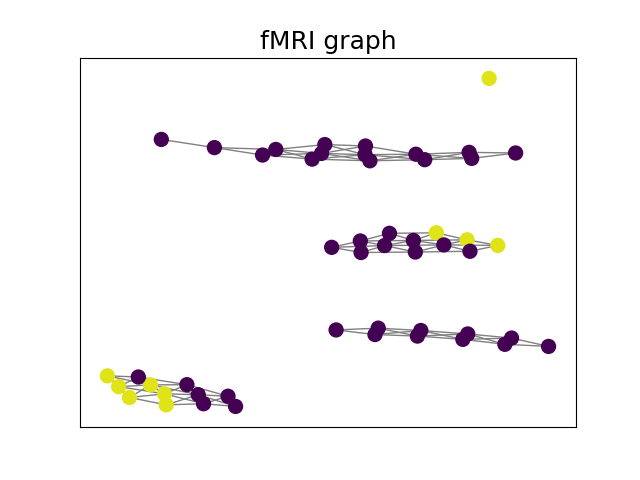}
\caption{}\label{fmri1}
\end{subfigure}
\begin{subfigure}[b]{0.32\linewidth}
\includegraphics[width=\linewidth, height = 3.8cm]{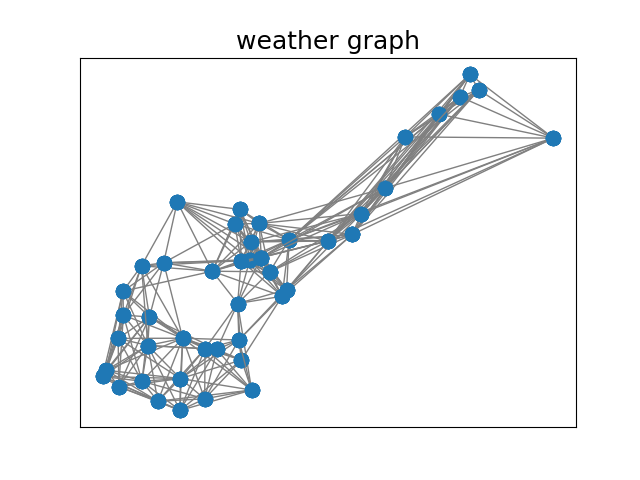}
\caption{}\label{weather1}
\end{subfigure}

\begin{subfigure}[b]{0.32\linewidth}
\includegraphics[width=\linewidth, height = 3.8cm]{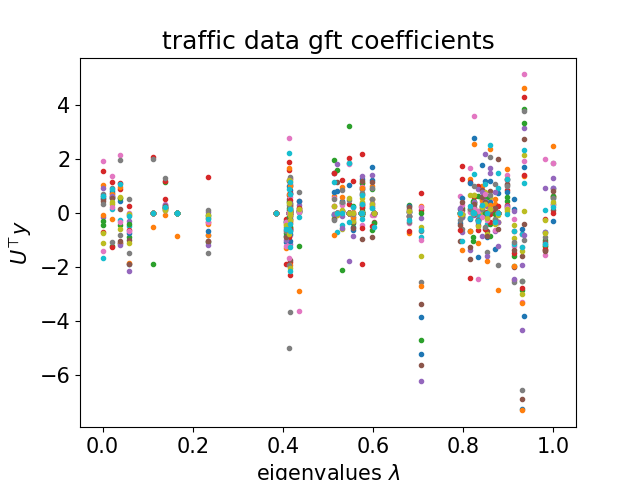}
\caption{}\label{traffic2}
\end{subfigure}
\begin{subfigure}[b]{0.32\linewidth}
\includegraphics[width=\linewidth, height = 3.8cm]{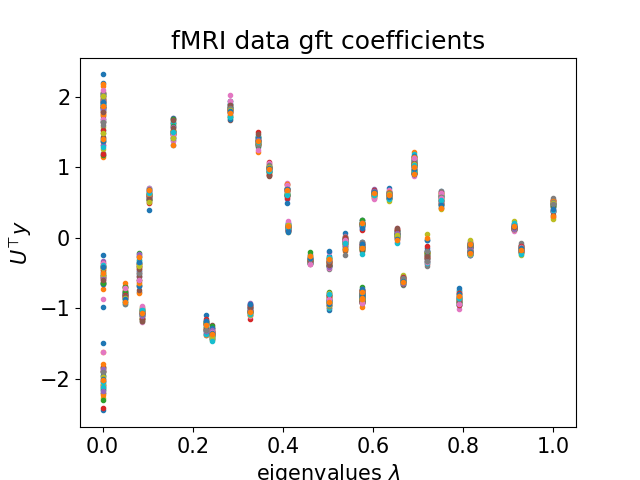}
\caption{}\label{fmri2}
\end{subfigure}
\begin{subfigure}[b]{0.32\linewidth}
\includegraphics[width=\linewidth, height = 3.8cm]{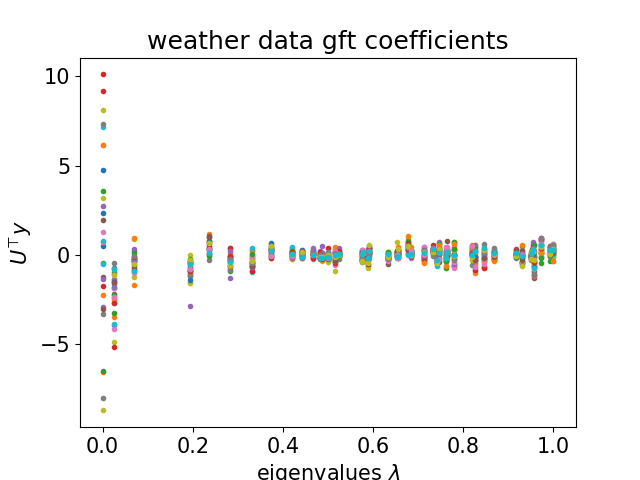}
\caption{}\label{weather2}
\end{subfigure}
\caption{Real world data graphs (a)-(c), where (a) and (b) show additionally how the graph is split into inputs (yellow) and outputs (purple). The graphs in (a) and (c) are generated using coordinates that correspond to physical locations, and the one in (b) using arbitrary coordinates. Plots (d)-(f) show the graph Fourier coefficients of the training data.} \label{info}
\end{figure}

We present in Fig. \ref{traffic1} to \ref{weather1} the graphs and the graph Fourier spectrum of the data in the three real-world experiments. In particular, for the traffic and fMRI data the graph is split into yellow and purple nodes, where the data observed on these nodes will correspond to inputs and outputs respectively for the GP. The signals we work with is on a graph that only consists of the purple nodes; all connections related to the input nodes are thereby ignored. On the weather dataset we carry out next-day prediction so each signal is on the full graph. The graph Fourier spectrums of the data are presented in Fig. \ref{traffic2} to \ref{weather2}, where the corresponding filters learnt are presented in the main text and can be found in Fig. 2.

\vfill

\end{document}